\documentclass[10pt,twocolumn,letterpaper]{article}
\usepackage{wacv}
\usepackage{booktabs}
\usepackage{times}
\usepackage{epsfig}
\usepackage{graphicx}
\usepackage{amsmath}
\usepackage{amssymb}
\usepackage{amsfonts}
\usepackage{textgreek}
\usepackage{algorithm}
\usepackage{algorithmicx}
\usepackage[noend]{algpseudocode}
\usepackage{colortbl}
\usepackage{arydshln}
\usepackage{multirow}
\usepackage{adjustbox}
\usepackage{url}
\usepackage{diagbox}
\usepackage{appendix}

\def\wacvPaperID{77} 

\wacvapplicationstrack 

\wacvfinalcopy 

\ifwacvfinal
\usepackage[breaklinks=true,bookmarks=false]{hyperref}
\else
\usepackage[pagebackref=true,breaklinks=true,colorlinks,bookmarks=false]{hyperref}
\fi

\begin{document}

\title{Analysis of Master Vein Attacks on Finger Vein Recognition Systems}

\author{Huy H. Nguyen$^1$, Trung-Nghia Le$^{1,3,4}$, Junichi Yamagishi$^{1}$, and Isao Echizen$^{1,2}$ \\
\small{$^1$National Institute of Informatics, Tokyo, Japan\ \ \ \ \ \ \ \ \ \ \ \ \ \ \ $^2$University of Tokyo, Tokyo, Japan} \\
\small{$^3$University of Science, VNU-HCM, Vietnam\ \ \ \ \ \ \ \ \ \ \ \ \ \ \ $^4$Vietnam National University, Ho Chi Minh City, Vietnam} \\
{\tt\small \{nhhuy,jyamagis,iechizen\}@nii.ac.jp}
}

\maketitle

\begin{abstract}
Finger vein recognition (FVR) systems have been commercially used, especially in ATMs, for customer verification. Thus, it is essential to measure their robustness against various attack methods, especially when a hand-crafted FVR system is used without any countermeasure methods. In this paper, we are the first in the literature to introduce master vein attacks in which we craft a vein-looking image so that it can falsely match with as many identities as possible by the FVR systems. We present two methods for generating master veins for use in attacking these systems. The first uses an adaptation of the latent variable evolution algorithm with a proposed generative model (a multi-stage combination of \textbeta-VAE and WGAN-GP models). The second uses an adversarial machine learning attack method to attack a strong surrogate CNN-based recognition system. The two methods can be easily combined to boost their attack ability. Experimental results demonstrated that the proposed methods alone and together achieved false acceptance rates up to 73.29\% and 88.79\%, respectively, against Miura's hand-crafted FVR system. We also point out that Miura's system is easily compromised by non-vein-looking samples generated by a WGAN-GP model with false acceptance rates up to 94.21\%. The results raise the alarm about the robustness of such systems and suggest that master vein attacks should be considered an important security measure.
\end{abstract}

\section{Introduction}
Finger vein authentication (using a FVR system~\cite{shaheed2018systematic}) was first commercially implemented in Japan in 1997 and has become well-recognized because of its application in ATMs to authenticate users~\cite{von2007biometric}. Its usage frees users from remembering and regularly changing passwords to maintain security. Due to their convenience, biometric authentication methods (including finger vein ones) have become widely used. Therefore, it is essential to evaluate their robustness and identify potential harms. A presentation attack is a common way to attack biometric recognition systems~\cite{marcel2019handbook}. Besides presenting a captured biometric trait of the victim, the attacker can use a wolf sample~\cite{une2007wolf}, which can match enrolled models of multiple identities. Master prints~\cite{bontrager2018deepmasterprints} and master faces~\cite{nguyen2020generating, shmelkin2021generating, nguyen2022masterface}  are examples of wolf samples generated using generative models. In reality, not all FVR systems have countermeasure methods deployed properly, allowing master vein attacks to compromise them.

\begin{figure*}[htb]
\centering
\includegraphics[width=140mm]{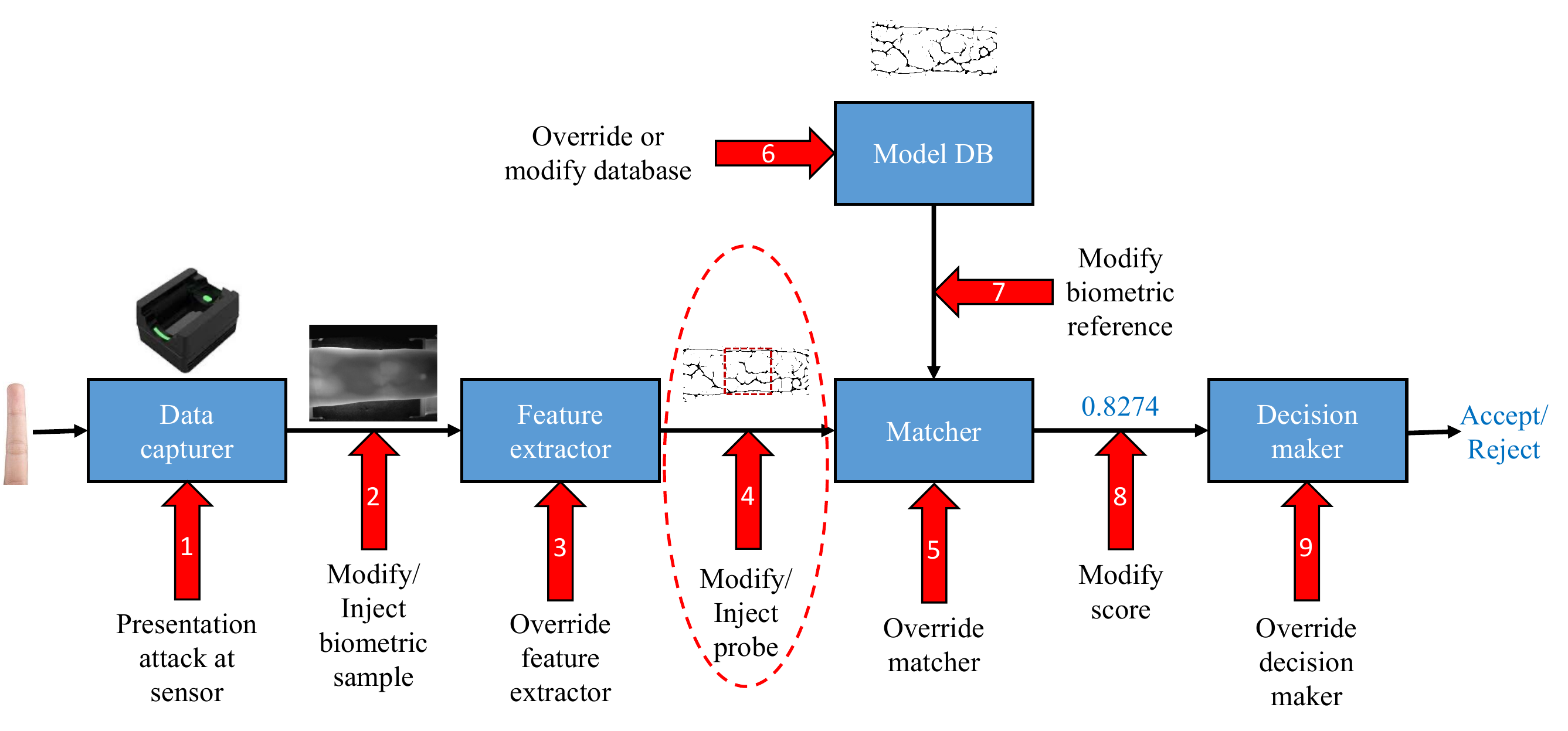}
\caption{Overview of FVR system and possible attacks on it, inspired by Ratha \textit{et al.}~\cite{ratha2001enhancing}. This paper focuses on attack 4 by injecting a master vein probe image.}
\label{fig:attacks}
\end{figure*}

Besides presentation attacks, there are several other ways that an attacker can compromise a biometric recognition system~\cite{ratha2001enhancing}, as shown in Fig.~\ref{fig:attacks}. Moreover, in theory, it is possible to craft a physical object (known as a presentation attack instrument, or PAI) from a synthetic master vein (an image clearly showing the center lines of the veins) and use it to perform a \textit{presentation attack} (attack 1 in Fig.~\ref{fig:attacks})~\cite{tome20151st}. It is also possible to translate the synthetic master vein into a captured vein sample using a convolutional neural network (CNN)~\cite{pang2021image} and use it to perform a \textit{logical attack} (attack 2 in Fig.~\ref{fig:attacks}). Due to these reasons, we focus on a logical attack with a clear vein image (attack 4 in Fig.~\ref{fig:attacks}) in this work. We craft a \textbf{master vein} image and then inject it as a probe to attack FVR systems built based on Miura \textit{et al.}'s design~\cite{miura2004feature, miura2007extraction}. \textit{A master vein image is a probe vein-looking image that can be falsely accepted as a match with enrolled models of multiple identities by a FVR system}. Although non-vein-looking images may have better attack ability, it is harder for them to generalize on other systems and other attack scenarios. There are two possible solutions to craft such master veins: using the latent variable evolution (LVE) algorithm~\cite{bontrager2018deepmasterprints} and using an adversarial machine learning (AdvML) attack~\cite{huang2011adversarial}.

The LVE algorithm is a common way to generate master biometric samples~\cite{bontrager2018deepmasterprints, nguyen2020generating, shmelkin2021generating, nguyen2022masterface}: a pre-trained generative model is used along with an evolutionary algorithm. The original work on generating master prints used a traditional generative adversarial network (GAN) model~\cite{gulrajani2017improved} while the work on generating master faces~\cite{nguyen2020generating, shmelkin2021generating, nguyen2022masterface} used an advanced GAN model trained on a large facial database~\cite{karras2019style}. With the original variational autoencoder (VAE)~\cite{kingma2014auto} and \textbeta-VAE~\cite{higgins2016beta, burgess2018understanding} generative models, there is a trade-off between image quality and the ability to learn disentangled representations. The traditional VAE and GAN~\cite{arjovsky2017wasserstein, gulrajani2017improved} models have trouble generating large images ($320 \times 240$ pixels in our case), while advanced models are data-hungry. The ability of the LVE algorithm to achieve good results depends on the disentanglement ability of the generative model. Moreover, compatibility with modern vein capturing devices depends on the model’s ability to generate high-resolution vein images.

An AdvML attack using an adversarial example can be used to change the output of a CNN~\cite{huang2011adversarial}. It is assumed that an attacker using a master vein attack does not know the identities of the enrolled models. The attacker thus attempts to generate a master vein that can match as many enrolled models as possible. State-of-the-art CNN-based FVR systems use different approaches between training and testing. For example, for a system~\cite{kuzu2021loss} that uses the additive angular margin loss~\cite{deng2019arcface} in training, the task is to minimize the cross-entropy loss or focal loss~\cite{lin2017focal} using the provided labels. In evaluation, cosine similarity is used to calculate the distance between the two embedded features of the probe and the model veins. Therefore, traditional adversarial attacks could not be applied in this case. Moreover, adversarial attacks are exclusive to machine-learning-based recognition systems. They are unlikely to generalize well to handcrafted recognition systems.

This work aims to solve to two above problems and then combine the two newly proposed solutions to generate master vein images that can attack both hand-crafted and deep-learning-based vein recognition systems. For the generative model used by the LVE algorithm, we proposed a method to combine the \textbeta-VAE model and the Wasserstein GAN with a gradient penalty (WGAN-GP)~\cite{gulrajani2017improved} model. The combination model can effectively learn disentanglement latent representations essential for the LVE algorithm and is capable of generating images with higher quality than the single models. Using this setting, we can successfully attack a hand-crafted system with about 70\% of false acceptance rates (FARs). However, this LVE-based method could not work on the CNN-based FVR systems, leading to the development of the adversarial-attack-based one. Unlike traditional adversarial attack methods, we propose using $k$ labels as targets. Since the target system uses cosine distance between the two embedded features in inference mode, we attack its training configuration, which uses an advanced addictive angular margin loss function. To make the attack more general, we combine these two proposed methods. By performing an adversarial attack on the master vein generated by the LVE-based method, the crafted master vein can fool both hand-crafted and CNN-based recognition systems with higher FARs (up to 88.79\% for the hand-crafted system) than those of master veins created by single methods.

In summary, the contributions of this work are four-fold:
\begin{itemize}
\item We point out that a hand-crafted vein recognition system without any countermeasure methods can be easily compromised by \textbf{non-vein-looking} images generated by a WGAN-GP model. We are also the first in the literature to investigate synthesized \textbf{vein-looking} images to perform master vein attacks.
\item We introduce a way to combine a \textbeta-VAE model and a WGAN-GP model to generate large, good-quality \textbf{vein-looking} images with better disentanglement. The trained \textbeta-VAE decoder extracted from this combination is then used in the LVE algorithm.
\item We present a $k$-label targeted adversarial attack for use in attacking a CNN-based FVR system. This target CNN-based system was trained using an advanced loss function (additive angular margin), outperforming a hand-crafted system.
\item We describe a highly successful attack that combines a latent LVE-based attack with an adversarial attack on a hand-crafted FVR system. We show that robustness against master vein attacks is an important measure for FVR systems.
\end{itemize}

\section{Related Work}
\subsection{Finger Vein Recognition Systems}
\label{sec:fvrs}
A typical vein recognition system usually has four modules (visualized in Fig.~\ref{fig:attacks}): a data capturer, a feature extractor, a matcher, and a decision maker~\cite{ratha2001enhancing}. Pre-processing operations may be applied before feature extraction. In the original work of Miura \textit{et al.}~\cite{miura2004feature, miura2007extraction}, the maximum curvature method and the repeated line tracking method were used for the feature extractor module and the cross-correlation method was used for the matcher module. The maximum curvature method was designed to be robust against varying vein widths and non-uniform brightness. We used it in a baseline handcrafted FVR system, which we refer to as ``Miura's system."

\begin{figure}[b!]
\centering
\includegraphics[width=60mm]{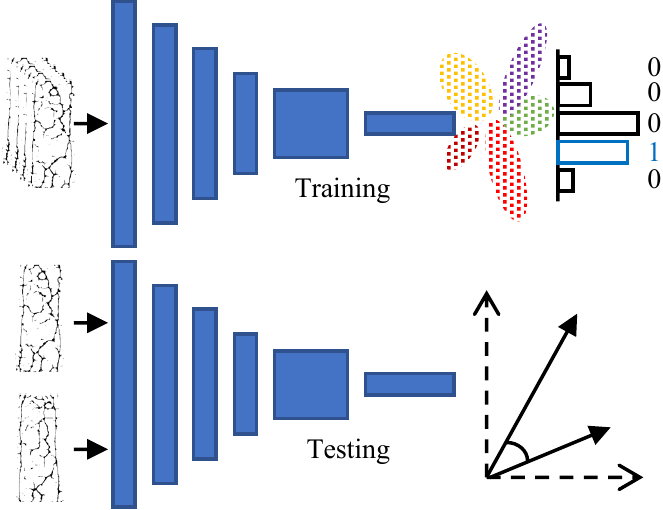}
\caption{Illustration of training and testing phases of CNN-based FVR system. The training phase uses the additive angular margin loss with ground-truth labels while the testing phase uses cosine distance between the two embedded features.}
\label{fig:fvrs}
\end{figure}

Besides fully handcrafted systems as introduced above, machine learning methods have been used to build the feature extractor and/or the matcher~\cite{shaheed2018systematic}. For instance, Kuzu \textit{et al.}~\cite{kuzu2021loss} used a modified version of the DenseNet-161 model~\cite{huang2017densely} to build a feature extractor and used cosine distance as the metric for matching. The modified DenseNet-161 model was trained using the additive angular margin loss~\cite{deng2019arcface} on the provided ground-truth labels. When testing, it was used to calculate the embedded features of the probe and the model finger vein. An overview of this kind of system is shown in Fig.~\ref{fig:fvrs}. To build a conceptual attackable CNN-based model for evaluation, we used this kind of CNN as an additional feature extractor to take the output of the maximum curvature-based feature extractor. We combined the additional CNN-based feature extractor with the cosine similarity-based matcher to form a new matcher. In our experiments, we use two modified versions of ResNet-18~\cite{he2016deep} and a modified version of MobileNetV3-Large~\cite{howard2019searching} (representing small networks) and ResNeXt-50~\cite{xie2017aggregated} (representing a large network) as the additional CNN feature extractors.

\subsection{Attacks on Biometric Recognition Systems}

A sample is considered a ``wolf'' if it can be falsely accepted as a match with models from multiple identities in a biometric recognition system~\cite{une2007wolf}. A wolf sample can be either biometric or non-biometric. Wolf attacks using wolf samples were initially used to target fingerprint recognition systems~\cite{ratha2001enhancing}. A master biometric attack is a particular case of a ``wolf attack'' in which the wolf sample looks similar to a biometric trait. A non-biometric wolf sample does not have any constraint on it, hence can be in any appearance. Therefore, non-biometric wolf samples are easier to craft and may have better attack ability than master biometric samples. However, a spoofing detector or a quality assessor integrated into a biometric recognition system can quickly reject them before being recognized. Moreover, since most non-biometric wolf samples mainly focus on a specific flaw in a particular system, they may not generalize well. Therefore, in this paper, we choose to investigate a master finger vein attack.

Master biometric attacks using master biometric samples have been recently used to attack partial fingerprint recognition systems~\cite{bontrager2018deepmasterprints} and face recognition systems~\cite{nguyen2020generating, shmelkin2021generating, nguyen2022masterface}. A latent variable evolution algorithm~\cite{bontrager2018deepmasterprints} is used to generate such master biometric traits by combining an evolutionary algorithm with a pre-trained generative model. The covariance matrix adaptation evolution strategy (CMA-ES)~\cite{hansen2001completely} is a popular choice for the evolution algorithm due to its novel design for non-linear and non-convex functions. It is sufficient for a low-resolution biometric trait (like partial fingerprints) generator to use a traditional generative model like WGAN-GP~\cite{gulrajani2017improved}. For high-resolution biometric traits like faces, it requires a more complex generative model like StyleGAN~\cite{karras2019style}. In this work, vein images do not need very high-resolution like facial ones, but low-resolution images like partial fingerprints are not sufficient. Thus, we could not simply use WGAN-GP as the generative model.

\section{Proposed Methods}
We first discuss the attack strategy used in this paper. We then introduce two methods for generating master veins, one using the LVE algorithm and one using an adversarial attack, and describe a way to combine them. We assume that the target FVR systems \textbf{do not use any spoofing detector or quality assessor}.

\subsection{Attack Strategy Analysis}
There are several positions where an attack can be carried out on a FVR system, as shown in Fig.~\ref{fig:attacks}. We aim to maximize the scope and effectiveness of master vein attacks given limited resources. It is crucial to ensure that the crafted master veins can be used to attack various systems under various conditions. The captured finger vein images are sensor-dependent, and the structure of the veins is unclear because of noise and lack of pre-processing. If we generate coarse master veins to carry out attack 2, the generative model cannot effectively learn the vein representations. The generated master veins also do not work well with other data capture devices. The vein images used to perform attack 4 are more precise, which is more suitable for training the generative model, performing attacks, and analysis. It is possible to translate a master vein into a corresponding captured image using a CNN~\cite{pang2021image} to perform attack 2. Moreover, an attacker can craft a corresponding PAI given an image of finger veins~\cite{tome20151st} that can be used to carry out a presentation attack (attack 1). In summary, \textit{in theory, it is possible to carry out attacks 1 and 2 if we can carry out attack 4}.

Miura's system can perform both symmetric matching (or full matching) and asymmetric matching (or partial matching). For partial matching, the probe is a randomly cropped image of the complete one. This is similar to the scenario in the work of Bontrager \textit{et al.}~\cite{bontrager2018deepmasterprints}. Before performing random cropping on an input vein image, the system uses an algorithm to calculate a mask to extract the vein-only region first. \textit{For simplicity, we assume that this region is provided. In reality, because of this algorithm, non-vein-looking master vein images may not be cropped appropriately, reducing their attack ability.} For CNN-based systems, the networks can only perform full matching. Therefore, to ensure generalizability, we focus on generating full master vein images. Furthermore, the full master vein images can be cropped for partial matching in Miura's system.

\subsection{Attack Using LVE-Based Method}

\subsubsection{Method's Description}

The work of Bontrager \textit{et al.}~\cite{bontrager2018deepmasterprints} used a WGAN-GP~\cite{gulrajani2017improved} to generate partial fingerprints (hereafter \textbf{LVE\textsuperscript{1}}). However, a WGAN-GP is hard to train, especially with limited training data. A \textbeta-VAE is easier to train and could learn better disentangled representations (hereafter \textbf{LVE\textsuperscript{2}}). However, its generated images have low quality. Therefore, we fuse their strengths in our proposed generator and use the LVE algorithm ~\cite{bontrager2018deepmasterprints, nguyen2020generating} to generate master veins (hereafter \textbf{LVE\textsuperscript{3}}).

\begin{figure}[t!]
\centering
\includegraphics[width=\columnwidth]{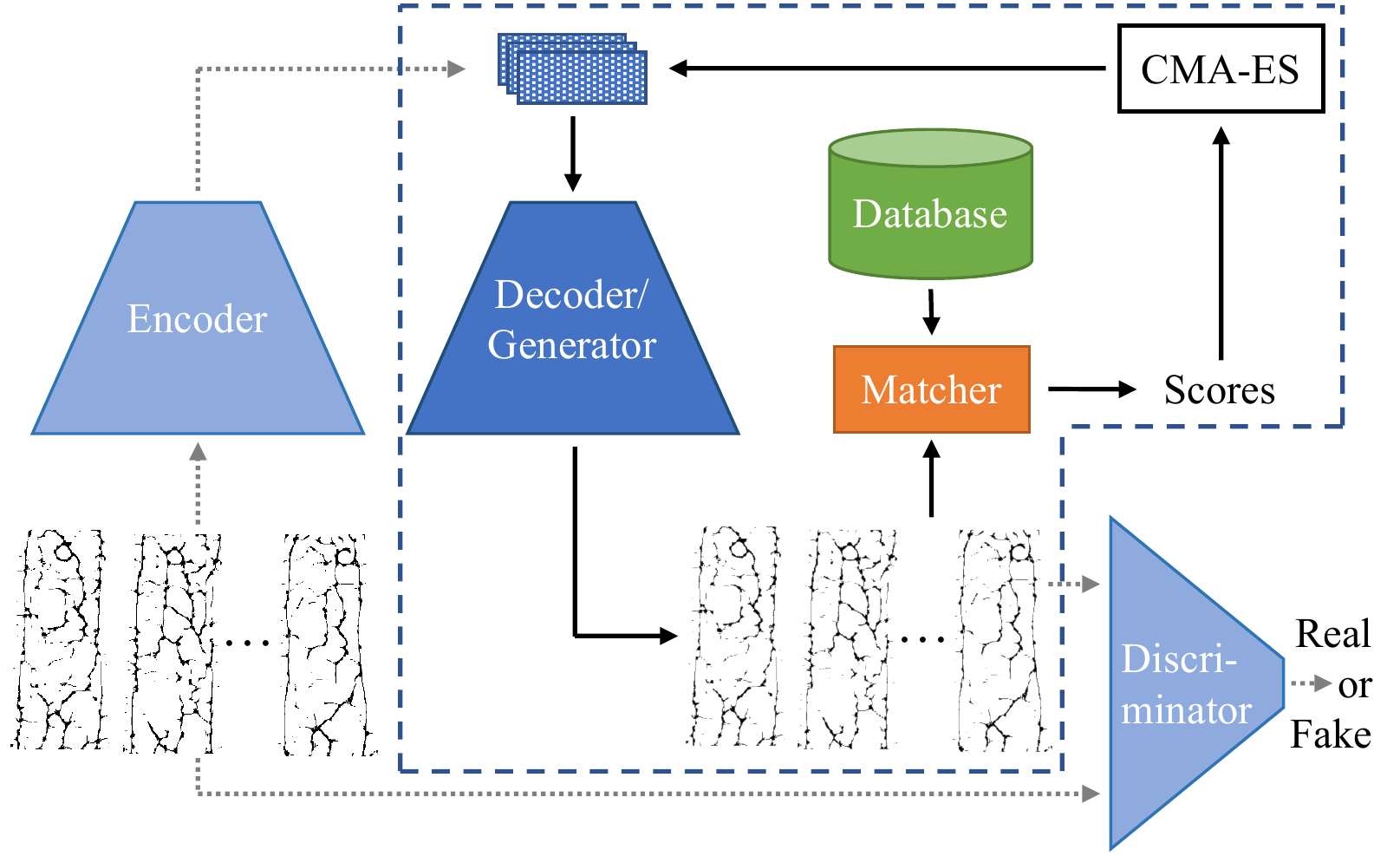}
\caption{Proposed LVE-based method. Only modules in the dashed polygon are used when running the LVE algorithm.}
\label{fig:lve}
\end{figure}

An overview of our proposed LVE\textsuperscript{3} method is shown in Fig.~\ref{fig:lve}. To achieve a better generative model with better learned disentanglement representations and high-resolution output, we first train a \textbeta-VAE model~\cite{higgins2016beta, burgess2018understanding} by minimizing Eq.~\ref{eq:bvae}.

\begin{equation}
\label{eq:bvae}
\begin{aligned}
    \mathcal{L}^{\text{\textbeta-VAE}}(\theta, \phi; \mathbf{x}, \mathbf{z}, C) = \mathbb{E}_{q_\phi(\mathbf{z}|\mathbf{x})}[\log p_\theta(\mathbf{x}|\mathbf{z})] \\
    - \gamma|D_{KL}(q_\phi(\mathbf{z}|{\mathbf{x}}) \parallel p_\theta(\mathbf{z})) - C| ,
\end{aligned}
\end{equation}
where $\phi$ and $\theta$ parameterize the distributions of the encoder $q_\phi$ and decoder $p_\theta$, respectively, and $D_{KL}(\ \parallel\ )$ represents Kullback-Leibler divergence.

\begin{figure}[t!]
\centering
\includegraphics[width=\columnwidth]{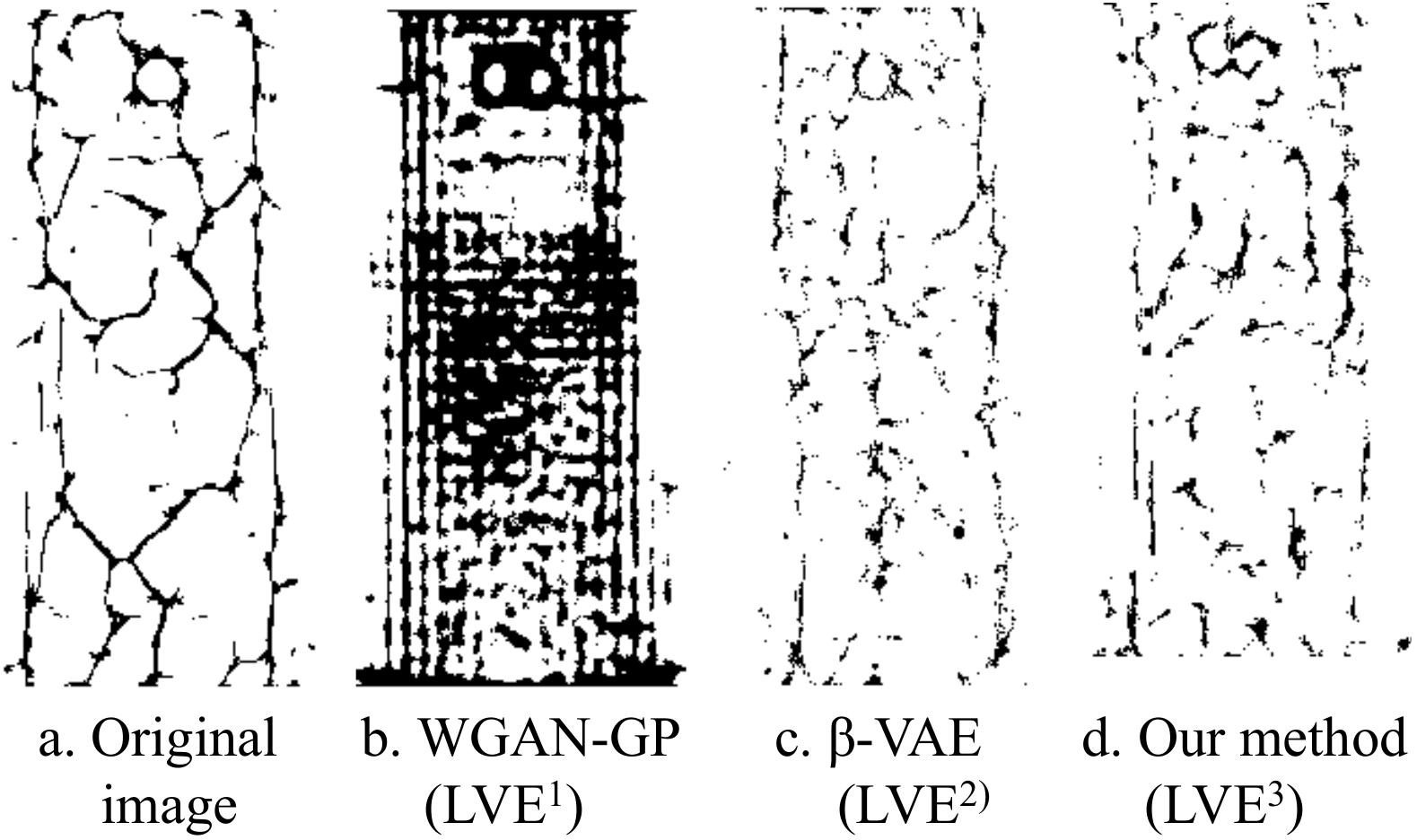}
\caption{Original image and images generated using WGAN-GP method, \textbeta-VAE method, and our proposed method (best viewed in the digital version with zoom-in). More samples are shown in the Supplementary Material. The WGAN-GP method failed to generate a vein-looking image while the \textbeta-VAE method generated a blurry image. Our proposed method generated a clearer image than the other two methods.}
\label{fig:gen_imgs}
\end{figure}

Then we fine-tune the decoder by using the WGAN-GP discriminator (minimizing Eq.~\ref{eq:wgangp}). Using this discriminator improves the quality of the generated images. To ensure stability, we freeze the parameters of $q_\phi$ and most of $p_\theta$ except for the last three convolutional layers of $p_\theta$ when minimizing $\mathcal{L}^{\text{GAN}}$. Finger vein images generated using the WGAN-GP method, the \textbeta-VAE method, and our proposed method are shown in Fig.~\ref{fig:gen_imgs} in this paper and in Fig.~\textcolor{red}{1} in the Supplementary Material. The WGAN-GP method failed to generate a realistic vein image while our method generated a clearer image than \textbeta-VAE.
\begin{equation}
\label{eq:wgangp}
\begin{aligned}
    \mathcal{L}^{\text{GAN}} = \mathbb{E}_{\tilde{\mathbf{x}} \sim
    \mathbb{P}_g}[D(\tilde{\mathbf{x}})] - \mathbb{E}_{\mathbf{x} \sim
    \mathbb{P}_r}[D(\mathbf{x})] \\
    + \lambda \mathbb{E}_{\hat{\mathbf{x}} \sim
    \mathbb{P}_{\hat{\mathbf{x}}}}[(\parallel \nabla_{\hat{\mathbf{x}}} D(\hat{\mathbf{x}}) \parallel_2 - 1)^2] ,
\end{aligned}
\end{equation}
where
\begin{itemize}
\item $\tilde{\mathbf{x}} = p_\theta(\mathbf{x}|\mathbf{z}) = p_\theta(\mathbf{x}|q_\phi(\mathbf{z}|\mathbf{x}))$.
\item $\mathbb{P}_r$ and $\mathbb{P}_g$ are the real and generated data distributions, respectively.
\item $\mathbb{P}_{\hat{\mathbf{x}}}$ is sampled uniformly along straight lines between pairs of points sampled from $\mathbb{P}_r$ and $\mathbb{P}_g$.
\end{itemize}

The LVE algorithm is described in Alg.~\textcolor{red}{1} in the Supplementary Material. For simplicity, we use the CMA-ES~\cite{hansen2001completely} for the evolutionary algorithm. Only the decoder $p_\theta$ of the \textbeta-VAE is used when running the LVE algorithm. It plays the role of the generator to generate vein images.

\subsubsection{Preliminary Analysis}
\label{sec:pre_analysis}
\begin{figure}[t!]
\centering
\includegraphics[width=\columnwidth]{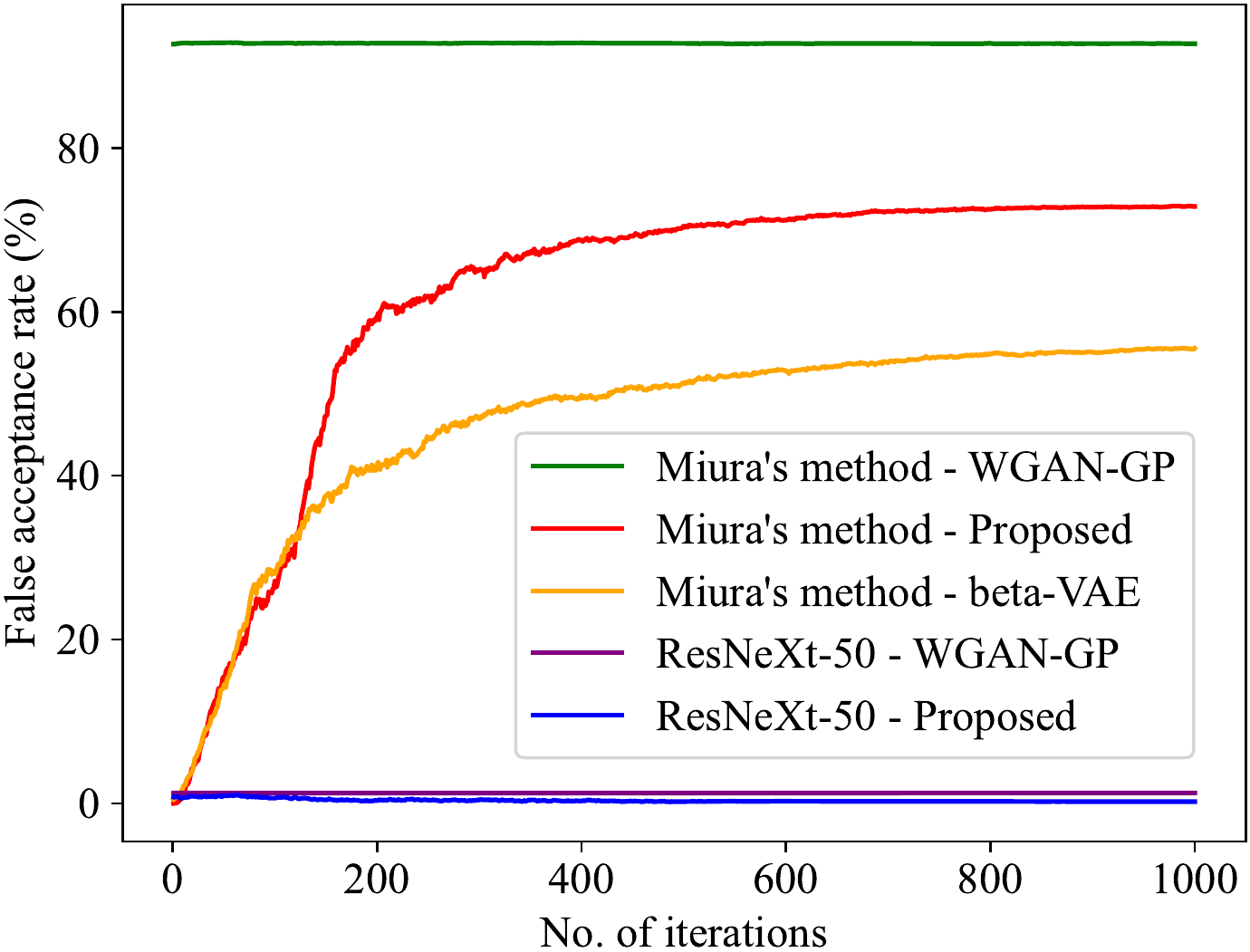}
\caption{FARs when running the LVE algorithm on Miura's and ResNeXt-50-based systems. When the number of iterations increases, only the FARs of Miura's system increase, implying that the LVE-based method only works on this system.}
\label{fig:far}
\end{figure}

The  false acceptance rate (FAR - the rate at which unauthorized or illegitimate users are verified) calculated when running the LVE algorithm are plotted in Fig.~\ref{fig:far}. The master vein generated using the LVE\textsuperscript{3} method on Miura's system is shown in Fig.~\ref{fig:masterveins}.b. Surprisingly, random non-vein-looking finger veins generated by the LVE\textsuperscript{1} method (using WGAN-GP) easily fooled Miura's system with the FARs higher than 90\%, even without the help of the LVE algorithm. Its cross-correlation-based matcher module failed to work correctly with these wolf samples. This finding raises \textbf{an urgent alarm on the reliability of the Miura's system without a spoofing detector or a quality assessor integrated}. 

Besides the above irregular case, the proposed LVE\textsuperscript{3} method worked better than the LVE\textsuperscript{2} method (using \textbeta-VAE) on Miura's system (with the FARs about 70\% and 50\%, respectively). This result confirms the effectiveness of our multi-stage combination of \textbeta-VAE and WGAN-GP for the generator. Although they are not perfect-looking, finger veins generated by the LVE\textsuperscript{3} method are more natural than those generated by the LVE\textsuperscript{1} and the LVE\textsuperscript{2} methods, reducing the possibility of being detected by the spoofing detectors or being rejected by the quality assessors. For a CNN-based recognition system, the LVE\textsuperscript{1} and LVE\textsuperscript{3} methods failed to work on the ResNeXt-50-based system with near-zero FARs. This failure may be due to the ResNeXt-50-based system being a large network trained on a well-designed loss function~\cite{deng2019arcface}, preventing the formation of dense clusters in its embedding space (discussed in Nguyen \textit{et al.}'s work~\cite{nguyen2022masterface}). We thus investigated another method to attack the CNN-based system, as described in the next section.

\subsection{Attack Using Adversarial Machine Learning Method}
We propose using a modified version of the $l_{\infty}$ projected gradient descent attack~\cite{madry2018towards} described by Eq.~\ref{eq:pgd}. We use a filter $K$ to control the shape of the perturbations, a mask $M$ to limit the area of the perturbations to the area containing the veins, and a soft-label vector $\mathbf{y}$ to control the target identities of the attack. Since the synthesized veins lay on fingers with similar shapes and locations, the mask $M$ can be easily approximated and crafted beforehand by hand or using a gap-filling algorithm.
\begin{equation}
\label{eq:pgd}
\begin{aligned}
\mathbf{x}^{t + 1} = \text{Clip}_{\mathbf{x}, \epsilon} (\mathbf{x}^t + \alpha (\zeta \ast K) \odot M) \\
\text{with}\ \ \ \zeta = \nabla_{\mathbf{x}}\mathcal{L}(\theta, \mathbf{x}^t, \mathbf{y}),
\end{aligned}
\end{equation}
where $\mathbf{x}$ is the input image, $\mathbf{y}$ is a target soft-label vector, $\theta$ is the set of target model parameters, $K$ is the filter kernel, $M$ is the finger vein mask, $\ast$ is the convolutional operator, and $\odot$ is the element-wise multiplication operator. 

\begin{figure}[t!]
\centering
\includegraphics[width=\linewidth]{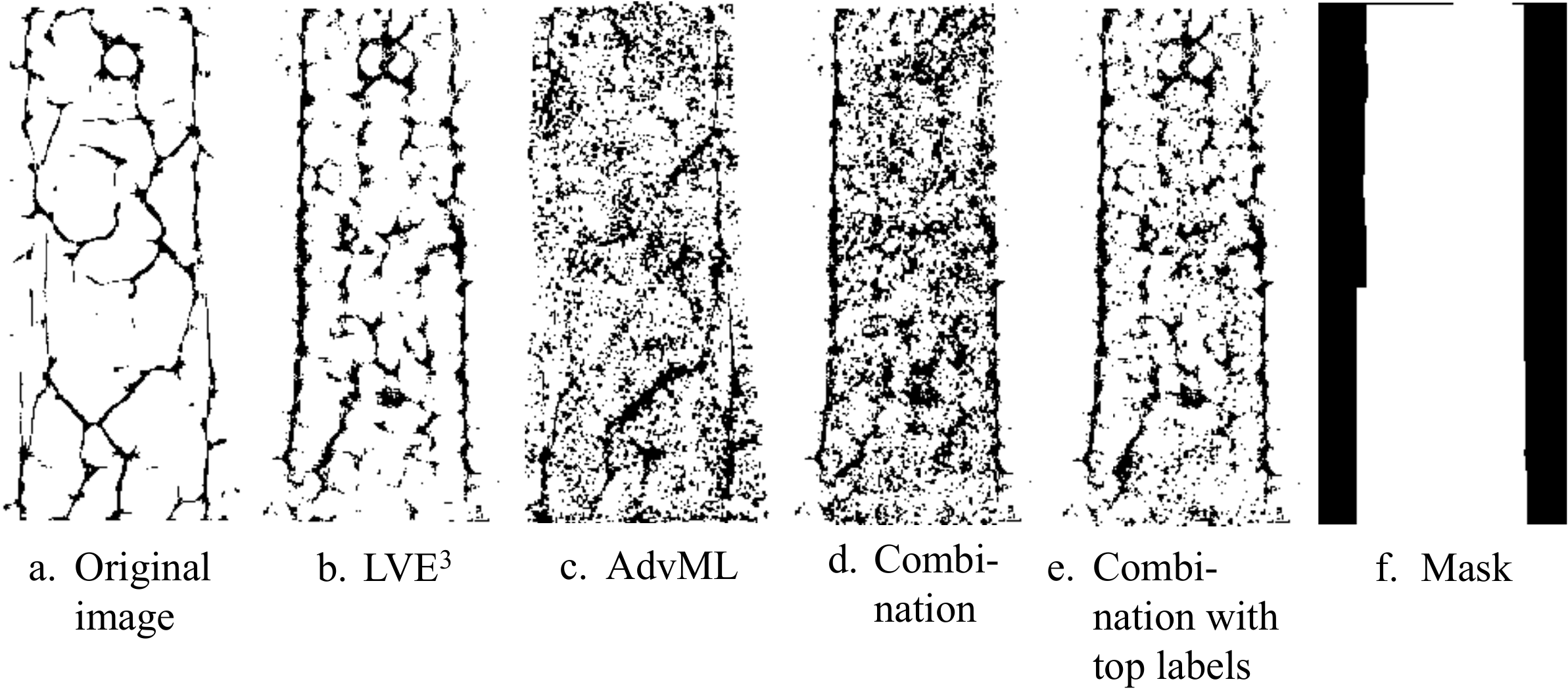}
\caption{Master veins generated using LVE method, adversarial machine learning method, their combination, and the corresponding mask used for adversarial attacks (best viewed in the digital version with zoom-in).}
\label{fig:masterveins}
\end{figure}

Unlike traditional adversarial targeted attacks, we target multiple labels instead of a single label. We call this \textbf{$k$-label targeted attack}. In more detail, we choose $1 < k < N$ of the total $N$ labels as target labels and set their probabilities to $1/k$, with $k$ as a hyper-parameter. For example, if we choose 3/5 of all labels, the target vector $\mathbf{y}$ is set to $[0.33, 0, 0.33, 0.33, 0]$. Since the FVR system does not calculate class probabilities during its testing phase but embeddings (see Fig.~\ref{fig:fvrs}), we need to attack the configuration of the training phase. In addition to randomly selecting $k$ target labels, we can choose the top-$k$ labels with the highest predicted probabilities. Doing so can make the optimizer process converge faster with fewer perturbations. Examples of randomly selected $k$ labels and top-$k$ labels are shown in Fig.~\ref{fig:masterveins}.d and Fig.~\ref{fig:masterveins}.e, respectively. If $k$ is close to 1, the attack is meaningless. Otherwise, it is hard to optimize the perturbations. In our experiments, the target CNN-based FVR system was powerful, with 0\% in both false acceptance rate and false rejection rate on the training set. Therefore, we used a small top-$k = 5\%$ in our experiments so that the optimization could successfully converge.

Besides the well-known hyper-parameter $\epsilon$, the number of iterations and the kind of filter $K$ are also important hyper-parameters. If the number of iterations is too small, the optimization will not converge. If it is too large, the perturbations will be too strong, damaging the original image. A damaged master vein image may be rejected by the quality assessor if implemented. Furthermore, it could not be generalized to other systems. Fig.~\textcolor{red}{2} in the Supplementary Material shows such an effect. When the number of iterations is around 200, the image is noisy, and when it is 500 or 1000, it is almost impossible to perceive the veins. We used 100 iterations in our experiments.

Regarding filter $K$, it is used to control the shape of the perturbations. Ideally, the perturbations should look like veins rather than random patterns or noise. In reality, it is very challenging to achieve this goal. We initially used a (differentiable) CNN-based vein/non-vein classifier as a loss function in optimizing the AdvML attack. However, it was easily fooled. On the other hand, a non-differentiable hand-crafted one was not useful for optimization with back-propagation. Therefore, we simply use a filter kernel for $K$ to regularize the perturbations. The goal is to avoid tiny-dot noise, which can be easily destroyed and is harder to adapt to other attack scenarios. We evaluated Gaussian blur, low-pass, high-pass, and Laplacian kernel. They have different effects on the rate of convergence of the optimization process and the quality of the master vein images. Examples are shown in Fig.~\textcolor{red}{3} in the Supplementary Material. The Gaussian blur kernel helped the optimization process converge the fastest and produced large-size perturbations with the least amount of tiny-dot noise (except the low-pass kernel), so it is the most suitable candidate kernel. The low-pass kernel destroyed almost all adversarial perturbations, preventing the adversarial attack. The high-pass and Laplacian kernels allowed too much tiny-dot noise. When a filter was not used, the optimizer process took a long time to converge, and the crafted perturbations were also tiny-dot noise.

Instead of using a bona fide image, we can use a master vein image as the input image $\mathbf{x}$. If we use a master vein image crafted using the LVE-based method, the corresponding adversarial image can work on both handcrafted recognition systems like Miura's one and CNN-based recognition systems, resulting in better generalizability. An example result of this combination attack is shown in Fig.~\ref{fig:masterveins}.

\section{Experiments}
We investigated the attack abilities of master veins crafted using the LVE-based method, the adversarial machine learning method, and their combination in white-box, gray-box, and black-box scenarios.

\subsection{Settings}
\subsubsection{Databases}
We used two finger vein databases:
\begin{itemize}
\item The SDUMLA-HMT Database~\cite{yin2011sdumla} contains images of six fingers per subject (six images per finger). We divided it into a training set containing the images for 80 subjects and a test set containing the images for 26 subjects. The training set was used to train the CNN-based recognition systems, set the recognition systems' matching thresholds, train the generative models, and generate master veins. We used both the training and test sets for evaluation.
\item The VERA FingerVein Database~\cite{tome20151st} contains bona fide images of two fingers of 110 subjects (two images per finger). We used the entire database to evaluate black-box attacks (in terms of database). Since its distribution is different from that of the SDUMLA-HMT Database, the recognition systems' matching thresholds calculated on the SDUMLA-HMT Database could not be used and needed to be re-set for this database.
\end{itemize}

\subsubsection{Finger Vein Recognition Systems}
We used three FVR systems: one hand-crafted system (Miura's system) and three CNN-based systems (one based on ResNet-18, one based on ResNeXt-50, and one based on MobileNetV3-Large). We chose MobileNetV3-Large (MobileNetV3-L) from the MobileNet family since it performed the best on FVR. For Miura's system (customized from \href{https://www.idiap.ch/software/bob/docs/bob/bob.bio.vein/stable/resources.html}{Idiap bob's implementation}), we evaluated both partial matching and full matching. We are aware of other hand-crafted recognition systems~\cite{shaheed2018systematic} using other features extracted from the local binary patterns, principal component analysis, or Gabor filters. This paper focuses on attack 4 in Fig.~\ref{fig:attacks}, however, these systems have different feature extractors. Attacking them requires building a mapping network to convert the master veins to their raw forms and perform attack number 2. We treated it as future work.

To generate master veins, we used Miura's system and the ResNeXt-50-based system as surrogate FVR systems. Since the LVE-based method failed to generate master veins on the ResNeXt-50-based recognition system, we ignored this case in all of our experiments. Hereafter, the LVE-based method is assumed to be the one running on Miura's system.

To evaluate performance, we used Miura's system and the ResNeXt-50-based system for evaluating white-box attacks and the ResNet-18- and  MobileNetV3-L-based ones for evaluating black-box attacks in terms of FVR systems.

\subsection{Evaluation Methodology}

We use FAR as the primary metric to define the effectiveness of master vein attacks. We compare the FARs of a FVR system on a normal dataset (without master veins) and a master vein dataset where the zero-effort imposter's probes were replaced by the master vein probes. If the FAR on the master vein dataset is moderately higher than the FAR on the normal dataset, the master vein attack is considered successful.

For the partial matching mode of Miura's system, we first randomly cropped the full vein images to the size of $128 \times 128$ pixels and then used them as the probes. Modal images are always full-size ones ($320 \times 240$ pixels).

\subsection{Results and Discussion}

\subsubsection{Attacks on Known Databases and Systems}

We first evaluated master vein attacks on the same and similar configurations we used to craft the master vein. In terms of databases, we attacked the SDUMLA-HMT one, resulting in a white-box attack. In terms of FVR systems, we attacked two types of systems, resulting in both white-box and black-box attacks. In more detail, attacks using master veins generated using Miura's system on CNN-based systems are black-box ones and vice versa, while attacking the same system are white-box ones. Merging both terms, we have white-box and gray-box attacks.

The FARs on bona fide veins and master veins are shown in Table~\ref{tab:FAR_standard}. Miura's system is extremely vulnerable to \textbf{non-vein-looking} wolf attack generated by the LVE\textsuperscript{1} method, with about 69\% for partial matching and 93.5\% for full matching, as mentioned earlier in section~\ref{sec:pre_analysis}. The possible reason is that when developing this system, wolf attacks had not been introduced. Therefore, the designation of its matching algorithm (based on cross-correlation) only considered vein-looking probes.

Miura's system is also vulnerable to \textbf{vein-looking} master vein attacks using the LVE\textsuperscript{2,3} methods, AdvML method, and their combination. The LVE\textsuperscript{3} method outperformed the LVE\textsuperscript{2} one and achieved the FARs of about 70\% on both train and test sets. It means that nearly two-thirds of the identities were falsely matched with the master veins. It happened because these generators can generate non-exist random finger vein images, and Miura's matcher algorithm has flaws. The LVE algorithm then guided them to generate finger vein images with wolf characteristics with large-enough iterations.

Although it is a black-box attack in terms of FVR systems, the AdvML method achieved the FARs of about 12\% for partial matching and about 40\% for full matching by Miura's system. The plausible explanation is that the AdvML master veins were optimized with full matching mode (CNN-based systems' only mode), their performance on partial matching mode is suboptimal.

A combination of the AdvML method and the LVE\textsuperscript{3} method sustainably raised the FARs of Miura's system in the full matching mode, which is about 85\%. However, it was not effective for the partial matching mode. Using the top probabilities label (denoted as (Top) in the table) helped increase the FARs on Miura's system in this partial matching mode and on the CNN-based systems. Its combination with the LVE\textsuperscript{1} method has the reverse tendency (possibly because LVE\textsuperscript{1} and LVE\textsuperscript{3} have different characteristics). Regarding robustness, CNN-based systems resist master vein attacks better than Miura's system. Their FARs only slightly increased (about 1 to 3\%) when attacks occurred. Regarding generalization, master vein attacks could not work well on unseen CNN-based systems (ResNet-18 and MobileNetV3-L).

\begin{table*}
\centering
\caption{FARs (in \%) of three FVR systems on SDUMLA-HMT Database with bona fide and master veins. Gray cells indicate gray-box attacks; white cells indicate white-box attacks. Bold numbers indicate that the master vein attacks have FARs higher than those of the corresponding bona fide cases by at least 1\%.} 
\label{tab:FAR_standard}
\begin{adjustbox}{max width=\textwidth}
\begin{tabular}{l|cc|cc|cc|cc|cc}
\multicolumn{1}{c|}{\textbf{Matcher}} & \multicolumn{2}{c|}{\begin{tabular}[c]{@{}c@{}}\textbf{Miura's system}\\\textbf{(Partial matching)}\end{tabular}} & \multicolumn{2}{c|}{\begin{tabular}[c]{@{}c@{}}\textbf{Miura's system}\\\textbf{(Full matching)}\end{tabular}} & \multicolumn{2}{c|}{\textbf{ResNeXt-50 }} & \multicolumn{2}{c|}{\textbf{\textbf{ResNet-18}}} & \multicolumn{2}{c}{\textbf{MobileNetV3-L}} \\ 
\hline
\multicolumn{1}{c|}{\textbf{Attack $\backslash$ Dataset}} & \textbf{Train set} & \textbf{Test set} & \textbf{Train set} & \textbf{Test set} & \textbf{Train set} & \textbf{Test set} & \textbf{\textbf{Train set}} & \textbf{\textbf{Test set}} & \multicolumn{1}{l}{\textbf{\textbf{\textbf{\textbf{Train set}}}}} & \multicolumn{1}{l}{\textbf{\textbf{\textbf{\textbf{Test set}}}}} \\ 
\hline
Bona fide & 07.57 & 08.02 & 08.46 & 08.98 & 0.00 & 2.25 & 0.00 & 3.37 & 0.00 & 1.31 \\ \hdashline[1pt/1pt]
LVE\textsuperscript{1} (WGAN-GP) & \textbf{68.24} & \textbf{70.41} & \textbf{92.46} & \textbf{94.21} &
{\cellcolor[rgb]{0.753,0.753,0.753}}\textbf{1.85} &
{\cellcolor[rgb]{0.753,0.753,0.753}}1.92 &
{\cellcolor[rgb]{0.753,0.753,0.753}}\textbf{1.51} &
{\cellcolor[rgb]{0.753,0.753,0.753}}2.25& 
{\cellcolor[rgb]{0.753,0.753,0.753}}0.67&
{\cellcolor[rgb]{0.753,0.753,0.753}}1.50\\
LVE\textsuperscript{2} (\textbeta-VAE) & \textbf{59.63} & \textbf{59.27} & \textbf{54.75} & \textbf{43.89} &
{\cellcolor[rgb]{0.753,0.753,0.753}}0.10 &
{\cellcolor[rgb]{0.753,0.753,0.753}}1.44 &
{\cellcolor[rgb]{0.753,0.753,0.753}}0.90 &
{\cellcolor[rgb]{0.753,0.753,0.753}}2.42 &
{\cellcolor[rgb]{0.753,0.753,0.753}}0.33 &
{\cellcolor[rgb]{0.753,0.753,0.753}}0.33 \\
LVE\textsuperscript{3} (Combination) & \textbf{70.47} & \textbf{69.85} & \textbf{73.29} & \textbf{71.84} & {\cellcolor[rgb]{0.753,0.753,0.753}}\textbf{1.46} & {\cellcolor[rgb]{0.753,0.753,0.753}}\textbf{6.07} & {\cellcolor[rgb]{0.753,0.753,0.753}}0.96 & {\cellcolor[rgb]{0.753,0.753,0.753}}\textbf{5.86} & {\cellcolor[rgb]{0.753,0.753,0.753}}0.53 &
{\cellcolor[rgb]{0.753,0.753,0.753}}2.03\\ \hdashline[1pt/1pt]
AdvML & {\cellcolor[rgb]{0.753,0.753,0.753}}\textbf{11.34} & {\cellcolor[rgb]{0.753,0.753,0.753}}\textbf{13.11} & {\cellcolor[rgb]{0.753,0.753,0.753}}\textbf{32.02} & {\cellcolor[rgb]{0.753,0.753,0.753}}\textbf{49.52} & \textbf{1.88} & \textbf{3.69} & {\cellcolor[rgb]{0.753,0.753,0.753}}\textbf{1.44} & {\cellcolor[rgb]{0.753,0.753,0.753}}2.24 &
{\cellcolor[rgb]{0.753,0.753,0.753}}0.61 &
{\cellcolor[rgb]{0.753,0.753,0.753}}1.46\\ \hdashline[1pt/1pt]
LVE\textsuperscript{3} + AdvML & \textbf{48.20} & \textbf{50.00} & \textbf{82.36} & \textbf{88.79} & \textbf{1.82} & \textbf{3.35} & {\cellcolor[rgb]{0.753,0.753,0.753}}\textbf{1.15} & {\cellcolor[rgb]{0.753,0.753,0.753}}1.93 & {\cellcolor[rgb]{0.753,0.753,0.753}}0.48 & {\cellcolor[rgb]{0.753,0.753,0.753}}0.64 \\
LVE\textsuperscript{3} + AdvML (Top) & \textbf{62.73} & \textbf{62.52} & \textbf{77.82} & \textbf{80.41} & \textbf{2.37} & \textbf{5.32} & {\cellcolor[rgb]{0.753,0.753,0.753}}\textbf{1.60} & {\cellcolor[rgb]{0.753,0.753,0.753}}4.00 & {\cellcolor[rgb]{0.753,0.753,0.753}}\textbf{1.03} & {\cellcolor[rgb]{0.753,0.753,0.753}}\textbf{3.47} \\ \hdashline[1pt/1pt]
LVE\textsuperscript{1} + AdvML (Top) & \textbf{76.60} & \textbf{76.95} & \textbf{91.86} & \textbf{93.81} & \textbf{1.68} & 1.85 &
{\cellcolor[rgb]{0.753,0.753,0.753}}\textbf{1.52} & {\cellcolor[rgb]{0.753,0.753,0.753}}2.09 & {\cellcolor[rgb]{0.753,0.753,0.753}}0.55 & {\cellcolor[rgb]{0.753,0.753,0.753}}0.40
\end{tabular}
\end{adjustbox}
\end{table*}

\begin{table}
\centering
\caption{FARs (in \%) of three FVR systems on VERA FingerVein Database with bona fide and master veins. Bold numbers indicate that the master vein attacks have FARs higher than those of the corresponding bona fide cases by at least 1\%.} 
\label{tab:FAR_blackbox}
\begin{adjustbox}{max width=\columnwidth}
\begin{tabular}{l|c:c:c:c:c}
\multicolumn{1}{c|}{\diagbox[width=2.2cm, height=1.6cm]{\textbf{Attack}}{\textbf{Matcher}}} & \begin{tabular}[c]{@{}c@{}}\textbf{Miura's}\\\textbf{system}\\\textbf{(Partial}\\\textbf{matching)}\end{tabular} & \begin{tabular}[c]{@{}c@{}}\textbf{\textbf{Miura's}}\\\textbf{\textbf{system}}\\\textbf{(Full}\\\textbf{matching)}\end{tabular} & \begin{tabular}[c]{@{}c@{}}\textbf{ResNeXt}\\\textbf{50}\end{tabular} & \begin{tabular}[c]{@{}c@{}}\textbf{ResNet}\\\textbf{18}\end{tabular} & \begin{tabular}[c]{@{}c@{}}\textbf{Mobile}\\\textbf{NetV3-L}\end{tabular} \\ 
\hline
Bona fide & 04.07 & 03.13 & 8.22 & 7.28 & 8.10 \\ \hdashline[1pt/1pt]
LVE\textsuperscript{1} (WGAN) & \textbf{38.84} & \textbf{43.86} & 0.18 & 0.10 & 0.18 \\
LVE\textsuperscript{2} (\textbeta-VAE) & \textbf{15.08} & 02.92 & 0.00 & 0.00 & 0.00 \\
LVE\textsuperscript{3} (Comb.) & \textbf{20.84} & \textbf{19.54} & 0.54 & 0.00 & 0.01 \\ \hdashline[1pt/1pt]
AdvML (A) & 03.12 & 03.57 & 0.20 & 0.04 &  0.18\\ \hdashline[1pt/1pt]
LVE\textsuperscript{3}+A & \textbf{16.37} & \textbf{47.73} & 0.42 & 0.01 &  0.18 \\
LVE\textsuperscript{3}+A (Top) & \textbf{22.25} & \textbf{26.34} & 0.82 & 0.52 & 0.21 \\ \hdashline[1pt/1pt]
LVE\textsuperscript{1}+A (Top) & \textbf{39.28} & \textbf{44.49} & 0.18 & 0.01 & 0.17 
\end{tabular}
\end{adjustbox}
\end{table}

\subsubsection{Cross-Database and Cross-System Attacks}

Next, we evaluated master vein attacks on more challenging scenarios. In terms of database, we attack a different database - the VERA FingerVein Database. In terms of FVR systems, attacks on Miura's system and the ResNeXt-50 system are white-box while attacks on ResNet-18 and MobileNetV3-L are black-box. Table~\ref{tab:FAR_blackbox} shows the FARs on bona fide and master veins.

Miura's system continued to be vulnerable to wolf attacks and master veins attacks. For master vein attacks, the FARs were around 20\% and could reach 47.73\% when we used the combination method to attack the full matching mode. Using top labels helped increase the FAR of the attack on partial matching mode to 22.25\%. On the other hand, the CNN-based recognition systems were robust against master vein attacks. However, it is important to note that the CNN-based recognition systems could not generalize well on the VERA FingerVein Database, resulting in higher FARs for bona fide vein attacks. It implies that if we want to use the CNN-based recognition systems effectively (so that bona fide users are not falsely rejected), we need to train them on the current dataset. However, doing so also opens a chance for master vein attacks.

\subsection{Summary}
Miura's system in partial matching and full-matching modes was vulnerable to non-vein-looking wolf attacks and vein-looking master vein attacks in white-box and gray-box scenarios. Both attacks substantially increased the FARs for Miura's system while barely increasing them for the CNN-based systems. A combination of the LVE\textsuperscript{3} method and the AdvML method can reach 88.79\% FAR on the full matching mode of Miura's system. Small increments of FARs on CNN-based systems indicate that they are more robust on master vein attacks.

It was challenging to perform master vein black-box attacks when both the target recognition system and the database were unknown. However, in reality, handcrafted FVR systems have already been deployed in ATMs, and the replacement cost is high. Since their variety is limited, attackers can narrow the scope of their attacks to gray-box or even white-box. Attackers can also prepare a set of potentially effective master veins in advance. Due to these reasons, master vein attacks still be a viable threat.

\subsection{Social Impacts}

To avoid possible harm, we use academic freely-accessed finger vein databases and open-source FVR systems. We believe our findings are necessary to raise awareness and promote improving the robustness of such systems. Besides robustness, we suggest using a fake finger vein detector to detect master veins.

\section{Conclusion and Future Work}

We have demonstrated that non-vein-looking wolf samples (generated by WGAN-GP) and vein-looking master veins generated using our proposed methods (LVE-based method, adversarial machine learning attack, and their combination) can successfully perform while-box and gray-box attacks on FVR systems. Miura's handcrafted system is fragile against such attacks, while CNN-based methods are more robust. Since not all commercial FVR systems are deep learning-based, their variety is limited, and the countermeasure methods are not always available, the threat of master vein attacks should not be underestimated.

Future work will focus on performing adversarial attacks to minimize cosine distance instead of maximizing label probabilities, improving the shape of adversarial perturbations to make them more vein-like, and evaluating more FVR systems and databases.

\section*{Acknowledgements}
This work was partially supported by JSPS KAKENHI Grants JP16H06302, JP18H04120, JP20K23355, JP21H04907, and JP21K18023, and by JST CREST Grants JPMJCR18A6 and JPMJCR20D3, including the AIP challenge program, Japan.

{\small
\bibliographystyle{ieee_fullname}
\bibliography{bibliography}
}

\onecolumn

\section*{Appendices}

\begin{appendices}

The Appendices are organized as follows. First, we provide the detail of the latent variable algorithm (LVE) used to craft master veins in section~\ref{sec:lve}. Next, we include some additional visualizations of real and generated finger veins in different settings in section~\ref{sec:visualization}. Last, we present an ablation study on top-$k$ AdvML attack in section~\ref{sec:albation}.

\section{Latent Variable Evolution Algorithm}
\label{sec:lve}

The LVE algorithm, which is used by the LVE-based methods to generate master veins (visualized in Fig.~\textcolor{red}{3} in the main paper), is described in detail by Alg.~\ref{alg:lve}. Regarding the implementation of the CMA-ES, we used the pycma library\footnote{\url{https://github.com/CMA-ES/pycma}}. For simplicity, we used its default parameters. 

\begin{algorithm*}[!htb]
	\caption{Latent variable evolution.}
	\label{alg:lve}
	\begin{algorithmic}
		\State $m \gets 18$
		\Comment{Population size, default value by the pycma library.}
		
		\Procedure{RunLVE}{$m, n$}
		\Comment{$m$ is population size and $n$ is total number of iterations.}
		\State MasterVeins = $\emptyset$
		\Comment{Master vein set.}
		\State Scores = $\emptyset$
		\Comment{and corresponding score set.}
		
		\State $\textbf{z} \gets \text{rand}()$
		\Comment{Initialize latent vectors $\mathbf{z} \in \mathbb{R}^{m}$.}
		
		\For {$n$ loops}
		\Comment{Run LVE algorithm $n$ times.}
		\State $\mathbf{V} \gets$ $p_\theta(\mathbf{z})$
		\Comment{Generate $m$ vein images $V$}
		
		\State $\mathbf{s} \gets 0$
		\Comment{Initialize scores $\mathbf{s} \in \mathbb{R}^{m}$.}
		
		\For {each vein image $V_i$ in $\mathbf{V}$}
		\For {each vein image $D_j$ in database $\mathbf{D}$}
		\State $s_i \gets s_i +$ Matcher($V_i, D_j$)
		\Comment{Calculate similarity score using matcher.}
		\EndFor
		
		\State $s_i \gets \frac{s_i}{|\mathbf{D}|}$
		\Comment{Calculate the mean scores.}
		
		\EndFor
		
		\State $V_b, s_b \gets$ GetBestVeinImage($\mathbf{V}, \mathbf{s}$)
		\Comment{Identify local best master vein image.}
		
		\State MasterVeins $\gets$ MasterVeins $\cup$ $\{V_b\}$
		\State Scores $\gets$ Scores $\cup$ $\{s_b\}$
		
		\State $\mathbf{z} \gets$ CMA\_ES($\mathbf{s}$)
		\Comment{Evolve $\mathbf{z}$ on basis of $\mathbf{s}$.}
		
		\EndFor
		
		\State  $V_{gb}, s_{gb} \gets$ GetBestVeinImage(MasterVeins, Scores)
		\Comment{Identify global best master vein image}
		
		\State \textbf{return} $V_{gb}, s_{gb}$
		\Comment{Best master vein and its score.}
		\EndProcedure
		
	\end{algorithmic}
\end{algorithm*}

\section{Additional Visualizations of Generated Finger Veins}
\label{sec:visualization}

\begin{figure*}[th!]
	\centering
	\includegraphics[width=\textwidth]{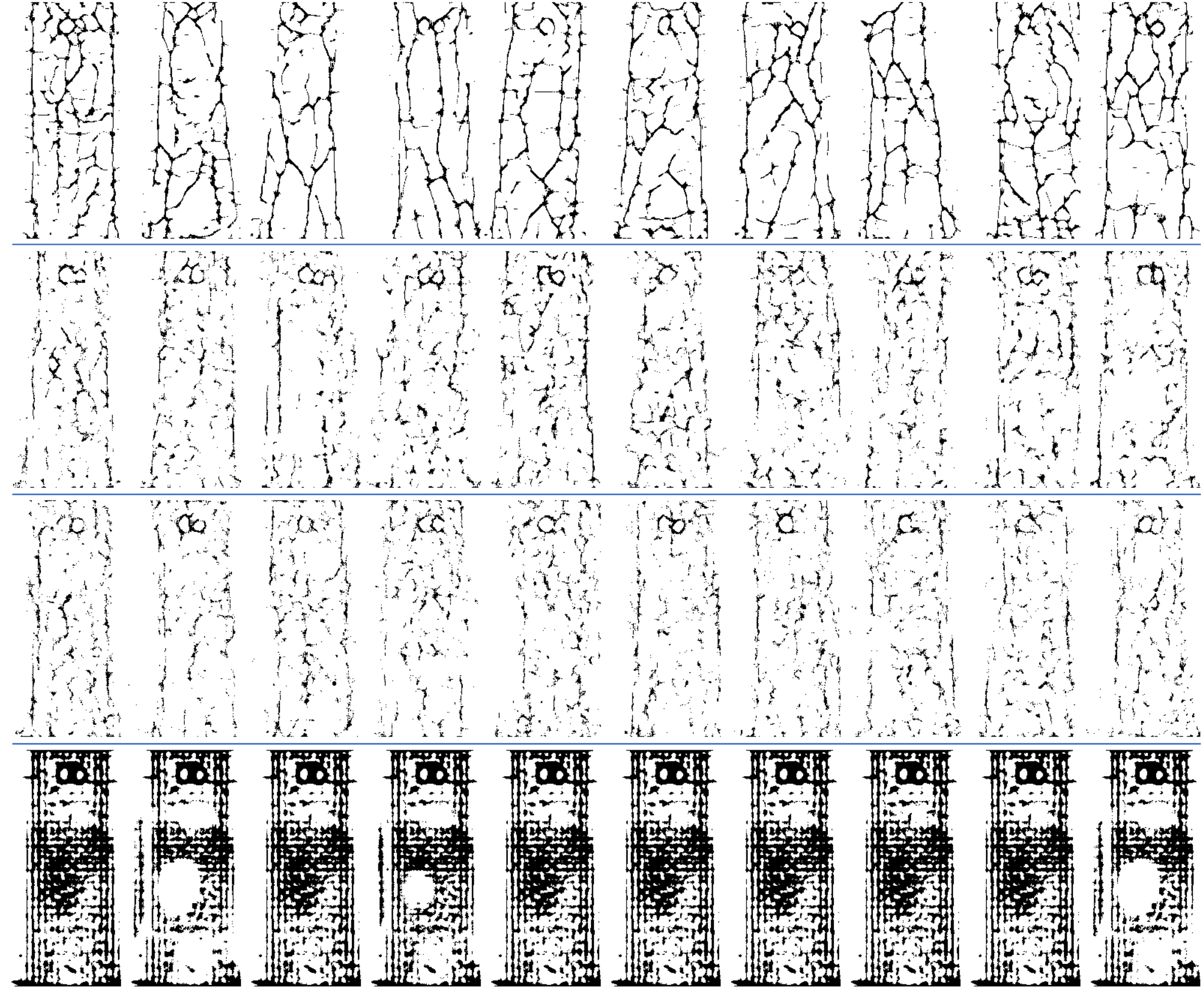}
	\caption{Examples of real finger veins (first row) and those generated by our proposed method (LVE\textsuperscript{3}, second row), \textbeta-VAE (LVE\textsuperscript{2}, third row), and WGAN-GP (LVE\textsuperscript{1}, last row). Since latent codes are sampling from noise, generated images have the randomness property. Among synthetic finger veins, those generated by our method had the best quality, while those generated by WGAN-GP do not look like finger veins.}
	\label{fig:samples_full}
\end{figure*}

\begin{figure*}[th!]
	\centering
	\includegraphics[width=120mm]{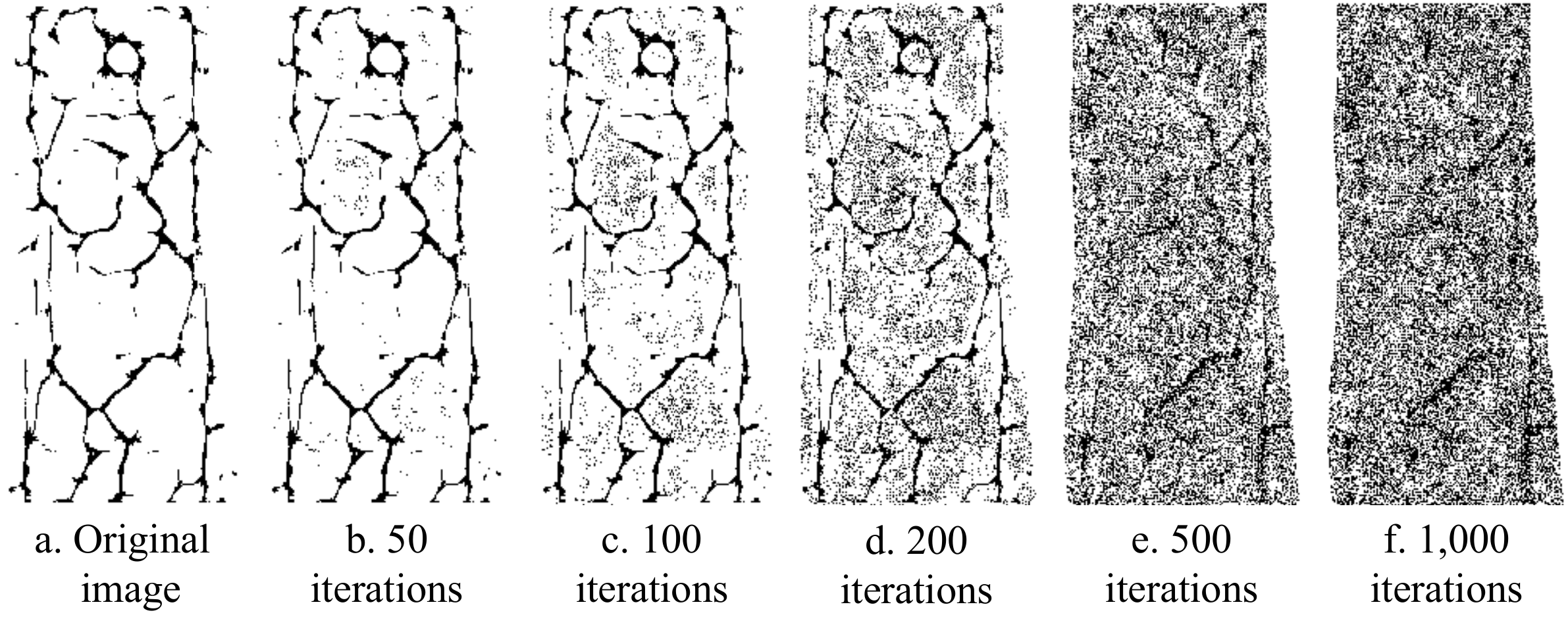}
	\caption{Effect of the number of iterations on the quality of adversarial master veins.}
	\label{fig:iteration}
\end{figure*}

\begin{figure*}[th!]
	\centering
	\includegraphics[width=120mm]{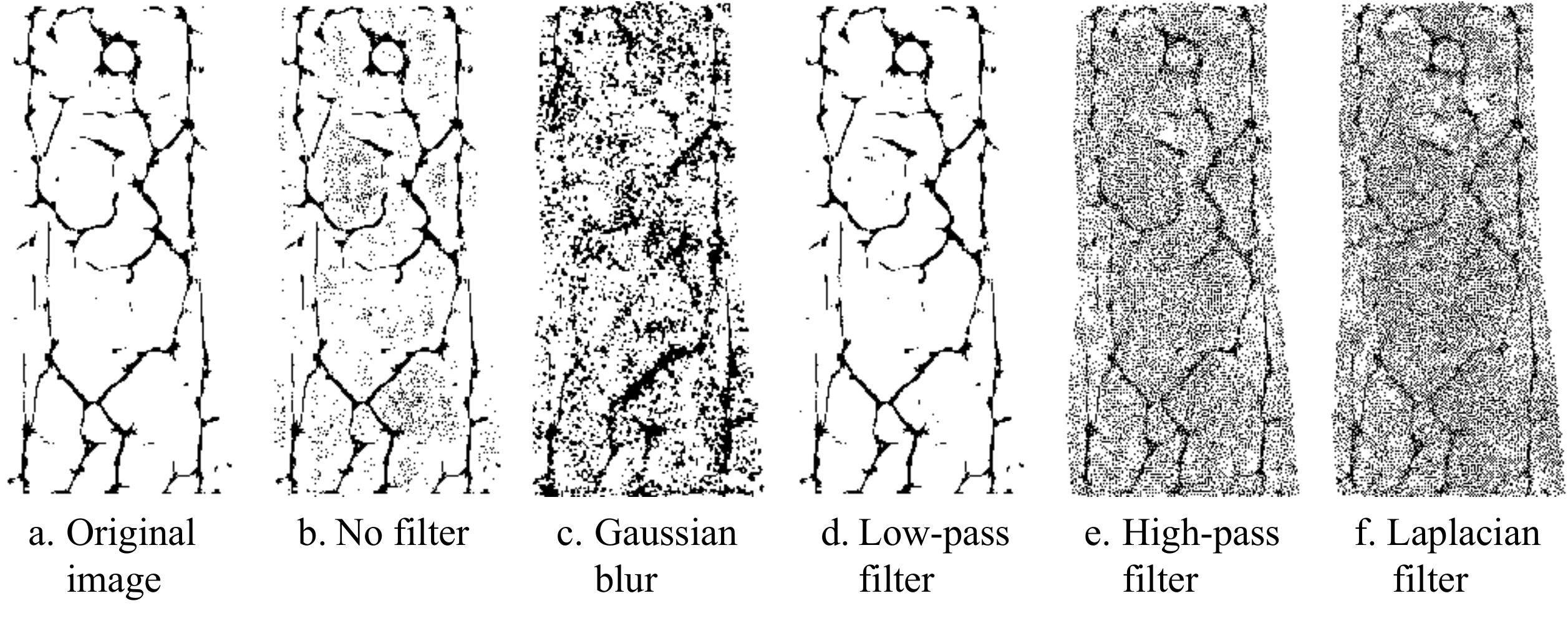}
	\caption{Effect of filters on the quality of adversarial master veins.}
	\label{fig:filters}
\end{figure*}

Additional samples of real finger veins and those generated by the LVE-based methods are shown in Fig.~\ref{fig:samples_full}. Effects of the number of iterations and filters used by the AdvML method on the quality of adversarial master veins are visualized in Figs.~\ref{fig:iteration} and~\ref{fig:filters}, respectively.

\begin{figure*}[th!]
	\centering
	\includegraphics[width=\textwidth]{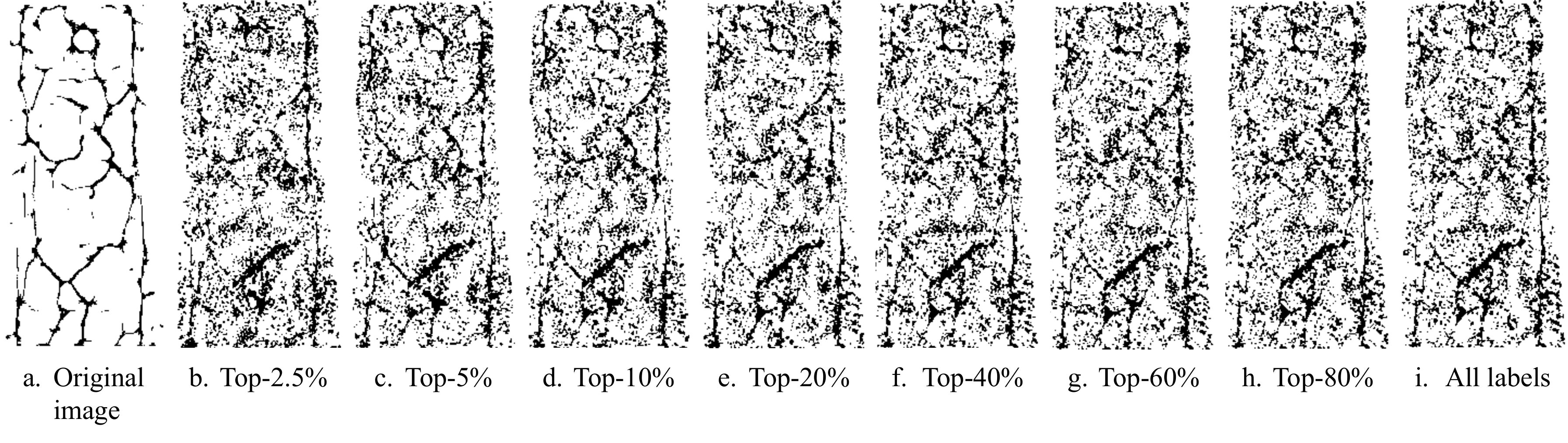}
	\caption{Adversarial master veins generated with different $k$ values in top-$k$-label targeted attack.}
	\label{fig:topk_veins}
\end{figure*}

\section{Ablation Study on Top-$k$ AdvML Attack}
\label{sec:albation}
\begin{figure*}[th!]
	\centering
	\includegraphics[width=90mm]{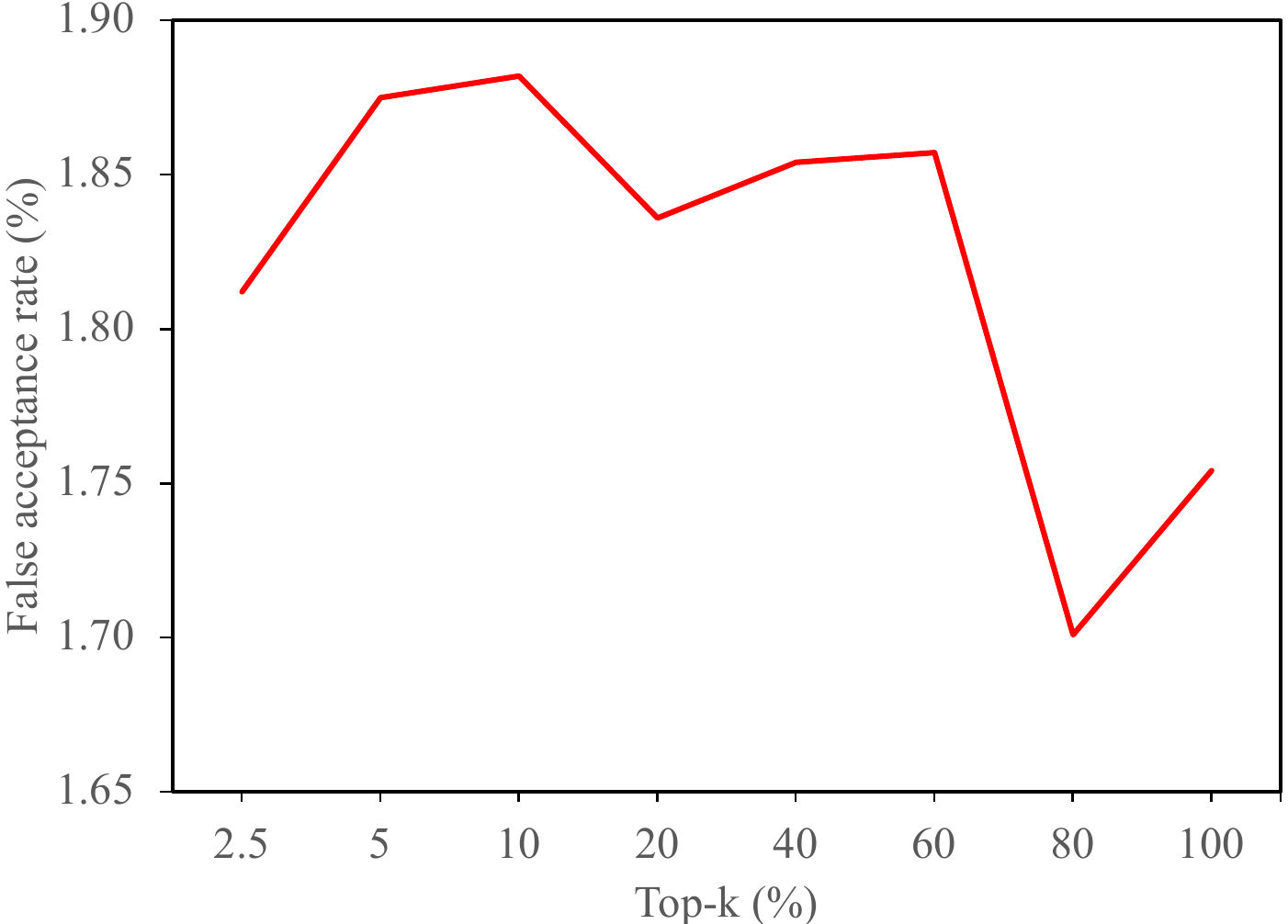}
	\caption{Relationship between $k$ and FARs in top-$k$-label targeted attack. FARs were calculated using the ResNeXt-50-based recognition system on the training set of the SDUMLA-HMT database.}
	\label{fig:topk_fars}
\end{figure*}

Examples of adversarial master veins generated with different $k$ values in the AdvML attack with top-$k$ labels are shown in Fig.~\ref{fig:topk_veins}. There are no significant differences in the amounts of perturbations between them. The relationship between $k$ and FARs is visualized in Fig.~\ref{fig:topk_fars}. There are no significant differences in the FARs, especially when $k$ is in the $[5\%, 60\%]$ range. In practice, it is better to avoid the extreme values of $k$. If $k$ is too small, its attack ability is limited. If $k$ is too large, it makes the optimization hard to converge.

\end{appendices}

\end{document}